%% file: arxiv.tex
\documentclass[10pt,twocolumn,letterpaper]{article}

\pdfoutput=1

\usepackage{3dv}
\usepackage{times}
\usepackage{epsfig}
\usepackage{graphicx}
\usepackage{amsmath}
\usepackage{amssymb}
\usepackage{multirow}
\usepackage{balance}
\usepackage{multicol}

\usepackage{hyperref}
\hypersetup{colorlinks,allcolors=green}

\threedvfinalcopy 

\def\threedvPaperID{2} 

\ifthreedvfinal\fi
\begin{document}

\title{Graphite: Graph-Induced feaTure Extraction for Point Cloud Registration}

\author{Mahdi Saleh$^{1}$, Shervin Dehghani$^{1}$, Benjamin Busam$^{1}$, Nassir Navab$^{1}$, Federico Tombari$^{1,2}$\\
$^{1}$ Technische Universit\"{a}t M\"{u}nchen,  $^{2}$ Google\\
{\tt\small \{m.saleh, shervin.dehghani, b.busam, nassir.navab\}@tum.de, tombari@in.tum.de}

}

\maketitle

\begin{abstract}
3D Point clouds are a rich source of information that enjoy growing popularity in the vision community. However, due to the sparsity of their representation, learning models based on large point clouds is still a challenge. In this work, we introduce \textit{Graphite}, a GRAPH-Induced feaTure Extraction pipeline, a simple yet powerful feature transform and keypoint detector. Graphite enables intensive down-sampling of point clouds with keypoint detection accompanied by a descriptor. We construct a generic graph-based learning scheme to describe point cloud regions and extract salient points. To this end, we take advantage of 6D pose information and metric learning to learn robust descriptions and keypoints across different scans. We Reformulate the 3D keypoint pipeline with graph neural networks which allow efficient processing of the point set while boosting its descriptive power which ultimately results in more accurate 3D registrations. We demonstrate our lightweight descriptor on common 3D descriptor matching and point cloud registration benchmarks \cite{Zeng3DMatch:Reconstructions,Wu3DShapes} and achieve comparable results with the state of the art. Describing 100 patches of a point cloud and detecting their keypoints takes only ~0.018 seconds with our proposed network.
\end{abstract}

\input{sections/Intro.tex}
\input{sections/Related.tex}
\input{sections/Method.tex}
\input{sections/Experiments.tex}
\input{sections/Conclusion.tex}

{\small
\bibliographystyle{ieee}
\bibliography{references}
}


\begin{center}
\onecolumn
\textbf{\large Supplemental Materials}
\end{center}
\begin{multicols}{2}

\setcounter{equation}{0}
\setcounter{figure}{0}
\setcounter{table}{0}
\setcounter{section}{0}
\renewcommand{\theequation}{S\arabic{equation}}
\renewcommand{\thefigure}{S\arabic{figure}}

In this supplementary material, we provide visualizations of all used datasets to enlighten how Graphite perceives point clouds. Moreover, we supply extended qualitative evaluations on the registration benchmarks for the interest of our readers.
\input{sections/supp/supp_1.tex}
\input{sections/supp/supp_2.tex}
\input{sections/supp/supp_3.tex}
\input{sections/supp/supp_4.tex}

\end{multicols}
\end{document}

%% file: sections/Intro.tex
\section{Introduction}
\begin{figure}
\centering
\includegraphics[width=\linewidth]{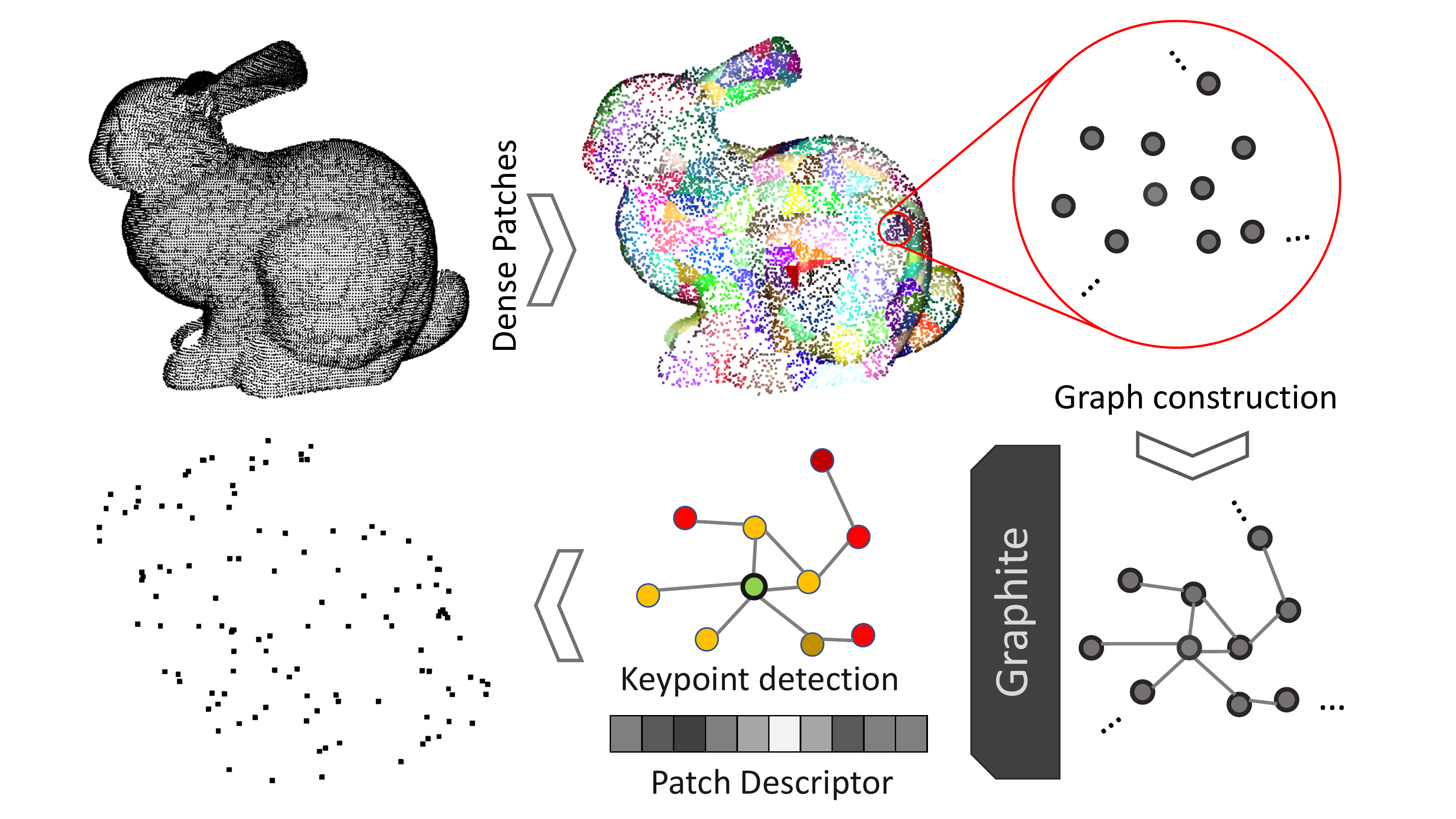}
\caption{Graphite helps representing a source point cloud (top-left) densely by describing each patch with a small descriptor and detecting a keypoint per patch (down-left). Fast calculation with lightweight model enables real-time point cloud applications. }
\label{fig:teaser}
\end{figure}
Point clouds play an indispensable role in modern 3D computer vision applications. LiDAR-equipped cars sense distances as sparse point clouds and mobile robots as well as modern hand-held devices measure depth with RGB-D and Time-of-Flight hardware. These devices essentially see the world in point cloud form. Thus, efficient processing of point clouds is an integral part to equip agents with 3D perception. At the same time, modern sensors are capable to provide large quantities of point cloud data at high frame rates. However, processing high numbers of sparse and unordered points can be computationally demanding. While trivial sub-sampling methods can lead to increased sparsity and loss of fine local information, perceptive sub-sampling of point clouds is very much structure dependant.\\
\noindent
\textbf{Keypoints and Feature Description. }
To distill the essential information in such rich vision data, classical methods represent image and scenes regions using feature descriptors. Extracted keypoints indicate salient landmarks and descriptors encapsulate local information in different formats. For example for the application of 6D pose estimation, a generic descriptor is oftentimes preferred over learning each object specifically with a deep neural network. In the image domain, handcrafted features \cite{Lowe2004DistinctiveKeypoints,bay2006surf,AlahiFREAK:Keypoint,Leutenegger2011BRISK:Keypoints} and learned descriptors \cite{Yi2016LIFT:Transform,verdie2015tilde} are employed to match regions or objects under certain levels of 3D rotation change and various illuminations. Recently with the popularity of self-supervised learning, it is possible to improve the quality of image-based features across multiple frames when homography or correspondence information is available. This has shown to improve matching rate when dense keypoints are exploited \cite{detone2018superpoint,DetoneTowardSLAM}. In the 3D domain and specifically on point clouds, descriptors are used for surface registration, multi-view reconstruction and 3D object pose estimation. Registration of 3D points clouds normally involves three steps: 1) find a set of salient points 2) describe each salient point with a descriptor 3) discover correspondences through descriptor matching.\\
\noindent
\textbf{Neural Point Processing. }
Point clouds are by nature unordered sets and processing models thus require permutation invariance. PointNet~\cite{qi2017pointnet} pioneered the field by suggesting a deep network that could segment or classify objects from their point clouds. In their work, building blocks such as T-Nets are utilized to force transformation invariance while permutation invariance is gained via max-pooling. Although PointNet and its successor PointNet++~\cite{Qi2017PointNet++:Space} are widely used, their design choice limits the inclusion of local neighborhoods and geometrical embeddings. Later works such as~\cite{NIPS2018_7362,Wang2019DynamicClouds} suggests the use of graphs to introduce point connectivity and vicinity information. Although they manage to learn local geometries by connecting points and learning node level features, they lack notion of metric scale through construction of graphs using k-NNs. After every step in processing node-level features, they change edge links and attributes. The inherent edge permutation of this process inhibits a consistent information flow. To this end, we propose a novel graph-based model capable of retaining geometrical shape under varying scale and sampling conditions. To describe point clouds, graphs are well suited as to capture relative information with their connections and can be designed to be invariant to transformations.\\
\noindent
\textbf{The Graphite Model. }
A classical approach for point cloud registration is the two stage paradigm of initial sampling of points to find keypoints and a subsequent feature description~\cite{RusuFastRegistration,GuoTriSI:Recognition}. A keypoint is matched with another by finding the closest point in feature space. If insufficient salient keypoints are matched or mismatches occur, the registration fails. Besides sub-sampling a point cloud to a set of interest points, we propose to describe point clouds on dense patches. However forcing keypoints per patch might not be ideal for some tasks such as registration. Therefore we define soft keypoint saliency through a scoring mechanism. We first form patches from a point cloud uniformly and then convert every patch to a graph structure similar to the connected structure of carbon atoms in the crystalline form of graphite. These graphs are learned to represent Graphite features which can be used for matching. We train our model in two stages. Firstly, we initialize our network to learn keypoints by supervision. In the second stage, we let our network optimize the descriptor and detector by performing metric learning on object point clouds obtained from various 6D poses.\\

\noindent
In short, our contribution is two-fold: Firstly, we provide a lightweight model that can detect salient keypoints in point cloud patches and regions taking advantage of a novel graph processing architecture. Secondly, the same model is used to describe the patch with a feature descriptor which can be used for correspondence matching. Our patch-based representation and reduction model is a major advantage compared to dense descriptors per point approach. We evaluate our descriptor on 3DMatch \cite{Zeng3DMatch:Reconstructions}, a commonly used benchmark for descriptor matching. Furthermore, we evaluate our dense descriptor and the extracted keypoints for the applications of point cloud registration on ModelNet Object Registration \cite{Wu3DShapes}.

%% file: sections/Related.tex
\section{Related Work}
This section reviews related work in the area of feature extraction in 2D to 3D from both handcrafted and learned approaches, and discusses state of the art for point cloud learning and registration. 

\subsection{2D Feature Extraction \& Matching}
Image descriptors are functions that map a local region of an image to a feature vector. In order to determine the interest region worth to describe, keypoint detectors are used to find salient areas. The first keypoint extractors \cite{moravec1977towards,HarrisADETECTOR} rely on image gradients and self-similarity metrics to detect corner like structures. To robustly detect corner points, consecutive methods make use of template-based techniques \cite{smith1997susan,rosten2006machine,rosten2008faster,mair2010adaptive} using machine learning and binary classifiers.
The concept of consecutively smoother images in scale space \cite{lindeberg1994scale,lindeberg1998feature} helps to find blob-structured keypoints across various image scales with the Laplacian of Gaussian (LoG) operator and its scale-normalized counterpart. The approximation of its calculation with a differences of Gaussians improves runtime performance in the prominent SIFT \cite{Lowe2004DistinctiveKeypoints} pipeline. Further methods \cite{bay2006surf,Leutenegger2011BRISK:Keypoints,AlahiFREAK:Keypoint,Lenc2016LearningDetectors} focus on advances in repeatability, accuracy, robustness and computational efficiency.
These are well studied in extensive comparison and survey papers \cite{schmid2000evaluation,tuytelaars2008local,mikolajczyk2005comparison,miksik2012evaluation,lee2014performance,salahat2017recent,BalntasHPatches:Descriptors} that investigate differences and performance of feature extraction pipelines.
With the advent of deep learning, TILDE \cite{verdie2015tilde} tackles the problem of illumination changes and MagicPoint \cite{detone2017toward} explores the advantages of training with synthetic primitive data while Key.Net \cite{Barroso-LagunaKey.Net:Filters} combines handcrafted and learnt features.\\
Some methods combine keypoint extraction directly with a feature description stage. SIFT, for instance, uses a 128-dimensional vector to describe its feature points and a ratio test is proposed to withdraw ambiguous matches. Further descriptors \cite{calonder2010brief,Leutenegger2011BRISK:Keypoints,Rublee2011ORB:SURF,AlahiFREAK:Keypoint} propose binary features to make use of the fact that the matching can be done efficiently via Hamming distance calculation. This allows their use in real-time systems such as 6D pose estimation \cite{busam2018markerless} and SLAM \cite{mur2017orb} pipelines.
Metric learning is a prominent way to leverage data for image description. HARDNet \cite{mishchuk2017working} proposes a triplet margin loss with hard negative mining for this task and the advantages of descriptor learning over hand-crafted methods have been shown with L2Net \cite{tian2017l2}. R2D2 \cite{r2d2} proposes a dense version of it while estimating also repeatability and reliability.

A joint estimation of detection, orientation and description is done in LIFT \cite{Yi2016LIFT:Transform} and LF-Net \cite{ono2018lf} also integrates scale space in an unsupervised learning framework driven by structure from motion. Similar to this, SuperPoint \cite{detone2018superpoint} uses homography warping as a self-supervision signal to jointly estimate saliency map and a dense descriptor. More recently, D2Net \cite{dusmanu2019d2} experiments with a single feature map for both detection and description resulting in more robust but less accurate results.
The matching stage classically involves a nearest neighbour search \cite{muja2009flann,beis1997shape,silpa2008optimised,fukunaga1975branch} accompanied by outlier removal through ratio test, cross check or RANSAC \cite{fischler1981random} and more elaborate techniques involve motion statistics \cite{bian2017gms} or temporal constraints \cite{ruhkamp2020dynamite}. Recently, SuperGlue \cite{SarlinSuperGlue:Networks} proposes to use a graph neural network to solve the assignment as an optimal transport problem.

\subsection{3D Descriptors}
3D descriptors are relatively less evolved. One reason is varied data representations and the complexity of describing point clouds. In scenarios where RGBD images are available, depth is used as an auxiliary information to find features or templates for matching \cite{HinterstoisserModelScenes,Kehl2016DeepEstimation,wohlhart2015learning}. Classical point cloud and surface descriptors such as SHOT \cite{tombari2010unique}, RoPS \cite{Guo2013RotationalRecognition} and TriSi \cite{GuoTriSI:Recognition} use a unique local reference frame to explain geometry with rotation invariance. PFH \cite{RusuAligningHistograms} and FPFH \cite{RusuFastRegistration} use pair-wise point features and surface normals describe curvature. While large scene variability can harm their performance, their classical nature allows for the use on edge devices with hardware constraints.
In recent years, scholars have also designed 3D descriptors with deep learning methods. CGF \cite{khoury2017learning} uses supervised learning to map hand-crafted high dimensional features into a lower dimensional vector. PPf-FoldNet \cite{Deng2018PPF-FoldNet:Descriptors} combines  PPF-Net \cite{Deng2018PPFNet:Matching}, PointNet \cite{qi2017pointnet} and FoldingNet\cite{Yang2018FoldingNet:Deformation} to learn rotation invariant features with self-supervision. 3DFeat-Net \cite{yew20183dfeat} also learns to extract sparse features with weak supervision from the tagged geolocation data. In parallel to this also dense voxel based approaches are explored. 3DMatch \cite{Zeng3DMatch:Reconstructions} converts point clouds to truncated distance functions (TDF) and Perfect Match \cite{GojcicTheDensities} uses voxelized smoothed density value (SDV) to describe local reference frames. Although they achieve substantial results on the 3DMatch Benchmark \cite{Zeng3DMatch:Reconstructions} they do not learn directly from point clouds.

\begin{figure*}[t]
\centering
\includegraphics[clip, trim=0.5cm 5cm 0.5cm 5cm,width=0.82\linewidth]{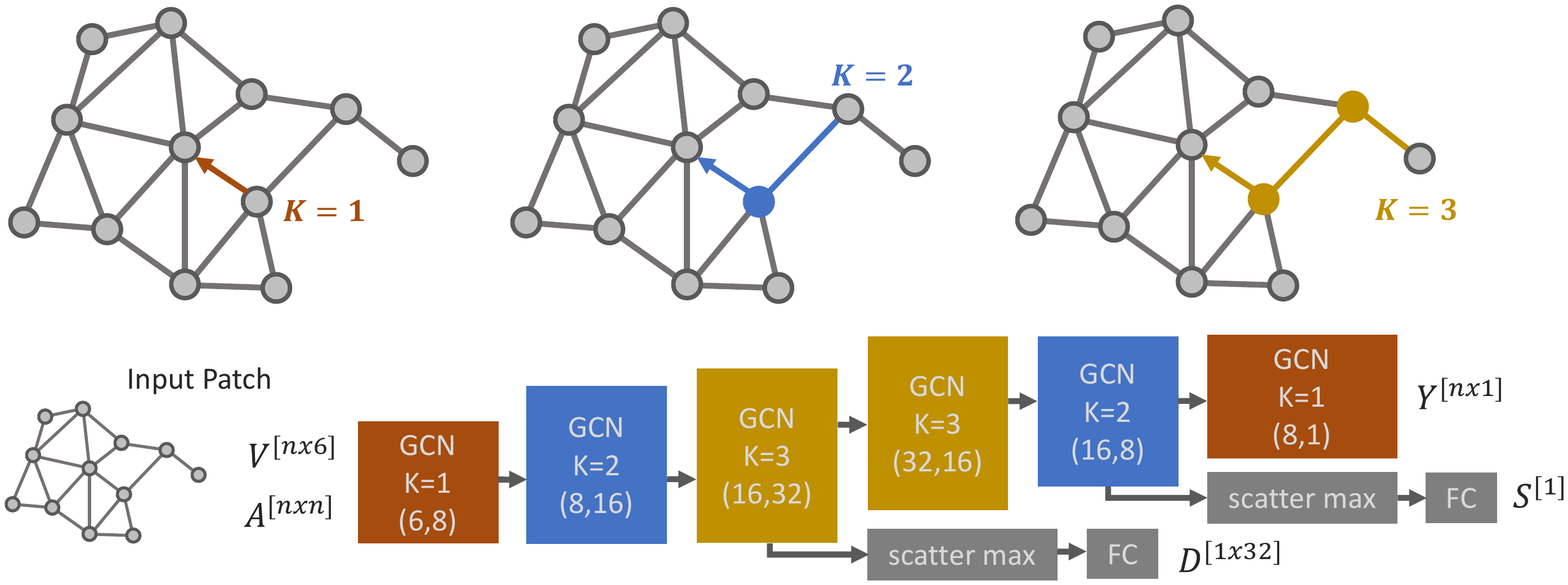}
\caption{Top: The schematic of node level feature propagation given the edge connectivites of graph and different hops values K=(1,2,3). Bottom: The Graphite architecture, consisting of GCN layers with different hops (increasing and then decreasing) to estimate a patch descriptor ($D$) from the middle stage and keypoint localization ($Y$) and scoring ($S$) at the end.}
\label{fig:network}
\end{figure*}

\subsection{Deep Point Cloud Processing}
While point clouds are used widely in computer vision, learning them with deep models have been far more challenging than 2D images or voxels. The pioneering pipeline PointNet \cite{qi2017pointnet} manages to build an architecture which is permutation-invariant and can learn point cloud features. Following that, PointNet++ \cite{Qi2017PointNet++:Space} proposes hierarchical learning to learn larger scale sets. While the contribution of PointNet is without doubt significant, it concentrates on global features. As a result, it fails to employ local features and information on the geometric neighborhood. To this end, also 3D Capsules \cite{zhao20193d} are used for this task where and auto-encoders and capsule networks learn point features. In an effort to simulate convolutions, researchers incorporated graphs to operate on point clouds \cite{MontiGeometricCNNs,Bronstein2017GeometricData}. Graphs help connecting points and build structures which can potentially represent surfaces and manifolds. Graph based point cloud approaches such as DGCNN \cite{Wang2019DynamicClouds} and PointCNN \cite{NIPS2018_7362} yield better results on segmentation and classification of point clouds. While graphs structures bring rotation invariance (isomorphism), how to define edges without losing metric information and attributing nodes and edges to sustain geometries are still not fully investigated.

\subsection{Point Cloud Registration}
Registering point clouds is a classic problem with applications such as scene reconstruction or object pose estimation. Conventional iterative methods such as ICP \cite{Arun1987Least-SquaresSets} are still commonly used although they are very dependent on initialization. Go-ICP \cite{Yang2016ARegistration} improves ICP by accuracy but at a high computational cost. Soft assignment \cite{Rangarajan1997TheAlgorithm} is another iterative approach which improves initialization by exploring and soft assigning correspondences to estimate 6D poses \cite{busam2015stereo}. Other scholars use conventional 3D descriptors to register clouds \cite{Tam2013RegistrationNonrigid,RusuFastRegistration,Yang2016ARegistration}. Following the success of PointNet, PointNetLK \cite{aoki2019pointnetlk} builds an iterative learning approach on top of PointNet and Lucas \& Kanade (LK) algorithm \cite{LucasAnVision} to register two point clouds. Deep Closest Point (DCP) \cite{wang2019deep} suggests using an attention-based module to find correspondences and a differentiable singular value decomposition to estimate transformation.


%% file: sections/Method.tex
\section{GRAPHITE}
In this section we explain the methodology and pipeline of Graphite. First we discuss how to segment point clouds into patches and to convert them into graphs. We then introduce our graph learning architecture and the training stage. Finally we explain how we perform point cloud warping.

\subsection{Point Patches}
\label{point_patches}
The input to our pipeline is a point cloud, that can come directly from a sensor such as a LiDAR, can be back-projected from an RGB-D frame or can be sampled from a 3D mesh. A point cloud $P$ with $m$ points is an unordered set of points $\{p_1,p_2,p_3,...,p_m\}$, with each $p_j$ consisting of coordinates and normal/color information $p_j = (x_j,y_j,z_j)$. We want to break our point cloud $P$ into small patches $C_i$ each holding $n$ points. $C_i = \{p_1,p_2,p_3,...,p_n\} $ with $\bigcup C_i = P$. In order to find clusters we perform random sampling and build clusters of $n$ points around each centroid. If a projected depth frame is available, this can be a rectangular patch of size $w \times w =m$ in the image domain, which represents a small frustum in the point cloud. Now suppose we have a function $F(C_i)=Y_i, F: {\rm I\!R}^{n \times 3} \rightarrow {\rm I\!R}^n$ that, for each point $p$ in patch $C_i$, finds a value $y$ representing the saliency of that point. The point with maximum value in each patch would be our point of interest or keypoint $k_i$:
\begin{equation} \label{eq1}
k_i=\underset{p}{\operatorname{argmax}}(F(C_i))=\underset{p}{\operatorname{argmax}}(Y_i)
\end{equation}

Every patch is thus associated with a salient point (keypoint). Furthermore, we transform every patch to a descriptor vector $D_i$ of size $l$. A function $G(C_i)=D_i, G: {\rm I\!R}^{n \times 3} \rightarrow {\rm I\!R}^{l}$ maps a patch to the descriptor. Using the functions $F$ and $G$, one can transform every patch to $k_i \in {\rm I\!R} ^ 3 $ and  $D_i \in {\rm I\!R} ^ l$ respectively.

\subsection{Graph Definition} \label{graphdef}
Given a set of points $C_i= \{p_1,p_2,p_3,...,p_n\}$ in a patch as described in the previous section, we want to create a graph $X_i=(V_i,E_i)$. Graphs are constructed from a set of vertices $V_i$ and edges $E_i$. As a common practice, we assume that every point in the patch $C_i$ can be considered as a node in the graph $X_i$. A node is represented by its coordinates and normal vectors in local reference frame $(x,y,z,a,b,c)$. Thus $V_i \in {\rm I\!R}^{n \times 6} $.

In addition to nodes, edges would demonstrate the connectivity of nodes and hold geometrical embedding. An edge $e_{j,k}$ connects vertices $v_j$ and $v_k$. In previous works such as \cite{Wang2019DynamicClouds}, edge connectivities are associated with K-nearest neighbors. The k-NN constraint would force a fixed number of edges into every node ultimately losing metric information. In contrast to this, we consider the unit ball with radius $r$ around each node's positional coordinates, and connect them to other nodes when their distance falls bellow $r$. We calculate the parameter $r$ based on average point cloud resolution. In addition to this, we attribute a weight to every edge to give a higher weight to close neighbors (and vice-versa) as follows:
\begin{equation}
\label{e_formula}
  e(j,k)=\begin{cases}
    \frac{r}{r+ ||p_j-p_k||}, & \text{if $||p_j-p_k||<r$}.\\
    0, & \text{otherwise}.
  \end{cases}
\end{equation}
Taking the function $F(C_i)$ from Section \ref{point_patches}, we can input a graph to our model $F(C_i)=F^\prime (X_i)=Y_i ,\; F^\prime: {\rm I\!R}^{n \times 6} \rightarrow {\rm I\!R}^n$ and predict a value per point/node of the patch. We can define this value as the inverse of the shortest path to the desired keypoint $k_i$. $y_{k_i}=max( F^\prime(X_i))$. During supervised training explained in Section \ref{init}, we assign the distance to a known keypoint to the node values.
\subsection{GCN Architecture}
In this section we introduce our graph convolutional network (GCN) architecture as depicted in Figure \ref{fig:network}. As mentioned previously we build graphs consisting of edges and nodes to represent a point cloud. We associate node features to estimate the value function $F(X)$ which is used for keypoint detection discussed in Section \ref{graphdef}. Moreover, we construct a function $G(X_i)$ that estimates descriptors from the graph, which should be invariant to node orders. We therefore use the scatter max operation which is a symmetric function for our descriptor. We build upon Topology Adaptive Graph Convolutions \cite{du2017topology} which will give us a framework to digest node and edge level information under different topologies.\\
We simulate the notion of multi-scale processing in classical feature descriptor and modern 2D convolutional networks by stacking GCN layers with different hops as illustrated in Figure \ref{fig:network}. Hops define how many nodes the information passes by on its way. By letting increasing and then decreasing hops (K=1,2,3) in our filters we increase the receptive field of nodes to capture patch-wide features and therefore manage to describe a global representation of the graph. Moreover, in contrast to previous works that uses two heads for $F$ and $G$ \cite{Barroso-LagunaKey.Net:Filters,detone2018superpoint}, we model both $F,G (X_i) = (Y_i,D_i)$ with the very same network. An illustration of hops and the Graphite model can be found in Figure \ref{fig:network}. In the next section, we describe the design and process steps of our three concepts for description, keypoint localization and scoring.
\subsection{Descriptor}
Starting with the input patch graph $X$ we apply 3 layers of graph convolutions at first. Every graph convolution module is taking into account the adjacency matrix $A \in {\rm I\!R}^{n \times n}$ built using Eq. \ref{e_formula}  and its diagonal degree matrix $D' \in {\rm I\!R}^{n \times n} $ to propagate node features across the graph. As in \cite{du2017topology}, we update node level information $X_i'$ by propagating features using edge level information as follows:
\begin{equation}
{X_i}^{\prime} = \sum_{k=0}^K \mathbf{D'}^{-1/2} \mathbf{A}^k
        \mathbf{D'}^{-1/2}\mathbf{X_i} \mathbf{\Theta}_{k},
\label{eq:tag}
\end{equation}
where ${\Theta}_{k}$ is the matrix to be learned for hop iteration $k$. After each layer we increase the depth of features from 6 to 8, 8 to 16 and 16 to 32. After three layers, the updated node features encapsulate features from the surrounding points and therefore describe the considered patch. To force descriptor invariance with respect to both point and node permutation, a scatter max aggregation function is utilized similar to PointNet \cite{qi2017pointnet}. The features follows linear fully connected layers and a normalization layer to project the feature on a unit sphere. The final descriptor vector is 32-dimensional ($l=32$).\\
\subsection{Keypoint Localization}
Apart from descriptor we need to formulate a value function $G(X_i)=Y_i$ to estimate how salient a point is in comparison with its neighboring nodes. In the context of this work we assume every small patch has at most one salient keypoint. Following the first three GCN layers leading to descriptor branch, we now squeeze back features to estimate per node values ($Y_i$). Feature depths are decreased from 32 to 16, 16 to 8 and 8 to 1 while the graph formation and node values are constrained.

\subsection{Keypoint Scoring}
\begin{figure}
\centering
\includegraphics[width=\linewidth]{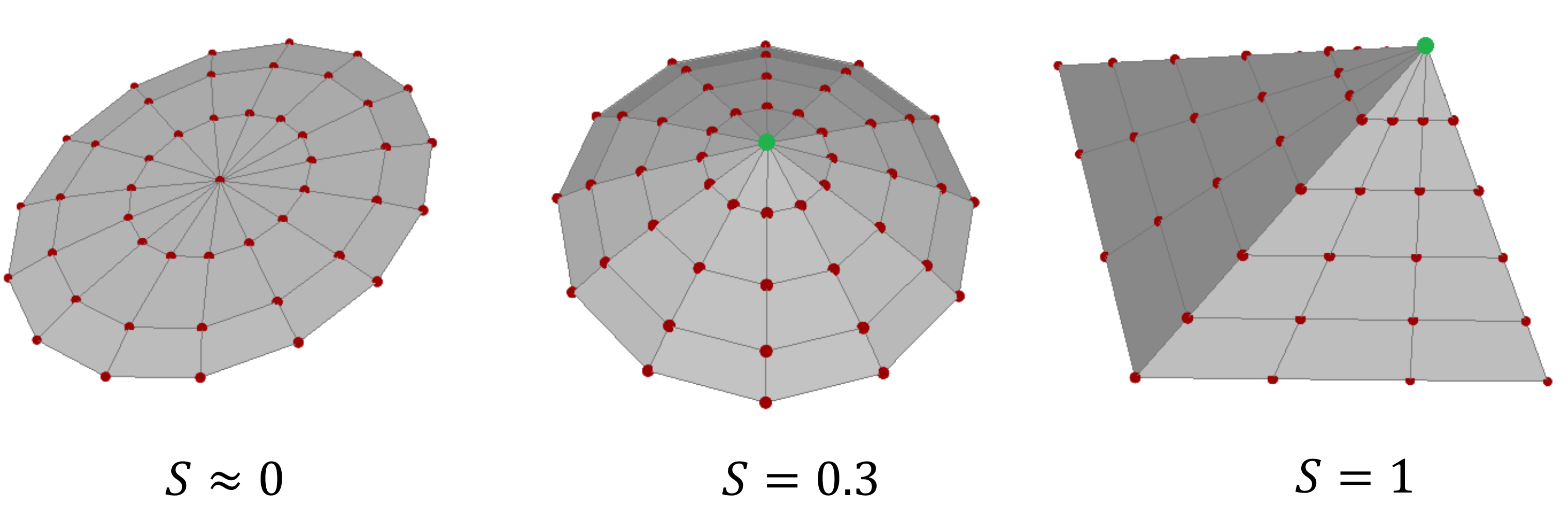}
\caption{We render point clouds (in red) from different surfaces of 3D primitives. We assign a score measure $S$ to each patch given its taper curvature and height and a keypoint $K$ in green. $S = 1$ infers a surface with potential salient keypoint . $S \approx 0$ infers a non-salient flat patch}
\label{fig:confidence}
\end{figure}

Our target is to improve point cloud registration. However, not every single patch possesses enough structure to represent a salient keypoint. For this reason, we formulate a soft invalidation of our non-salient keypoints through assignment of a global score $S$ per patch. Figure \ref{fig:confidence} depicts different level of keypoint scoring. We can use mesh surface information from synthetic data to assign scores given their curvature magnitude. To regress this scalar value we perform another scatter max on the output feature of layer five and apply a FC unit on top of that.

\subsection{Network Initialization} \label{init}

In order to properly learn our joint model we propose to have a training initialization. Supervising the network in the first stage helps the model to detect corners. As the pointiness of a shape corner is a fundamental geometric property, we let our network first focus on these points. Furthermore, we drive our GCN to predict surface flatness and curvature to infer keypoint scoring.

To fulfill this goal, we create a synthetic dataset, including point clouds of rendered depth maps with known shape and surface. Random primitive corners are generated with different curvature and known corner points. We also render the same 3d primitive (such as cone, box or pyramid) from two different camera views to simulate pose and sampling variations. Keypoint locations are labeled on the point sampled closest to the corner. We then measure ground truth values for all the points in the point cloud given their shortest path to the target keypoint in the constructed graph. Values and scores are learned in a supervised way with MSE loss (Eq. \ref{v_loss_formula}) to give the network an early assumption of such salient extreme points. This will be fine-tuned without direct supervision in the following stage (see Section \ref{posetraining}).
\begin{equation}
\label{v_loss_formula}
\mathcal{L}_V= (Y-\hat{Y})^2 ,\; \mathcal{L}_S= (S-\hat{S})^2
\end{equation}
Alongside keypoint detection, we also predict a descriptor which we train using metric learning with a triplet margin loss. Each of the triplet samples, \ie reference, positive and negative, will have a predicted descriptor ($D_r$, $D_p$ and $D_n$). A similarity term $|D_r-D_p|$ pulls together descriptors while a push term $|D_r-D_n|$ increases the distance between non-matching descriptors. The loss is as follows:
\begin{equation}
  \mathcal{L}_D= \frac{|D_r-D_p|}{|D_r-D_p|+m \cdot |D_r-D_n|} 
  \label{eq:triplet_loss}
  \end{equation}

With this we force the point clouds taken from the same patch but with different samplings to retrieve the same descriptor while descriptors from different patches are pushed apart. The total loss will accumulate all three losse terms $L_T=L_D+L_V+L_V$.

\begin{figure}
\centering
\includegraphics[width=\linewidth]{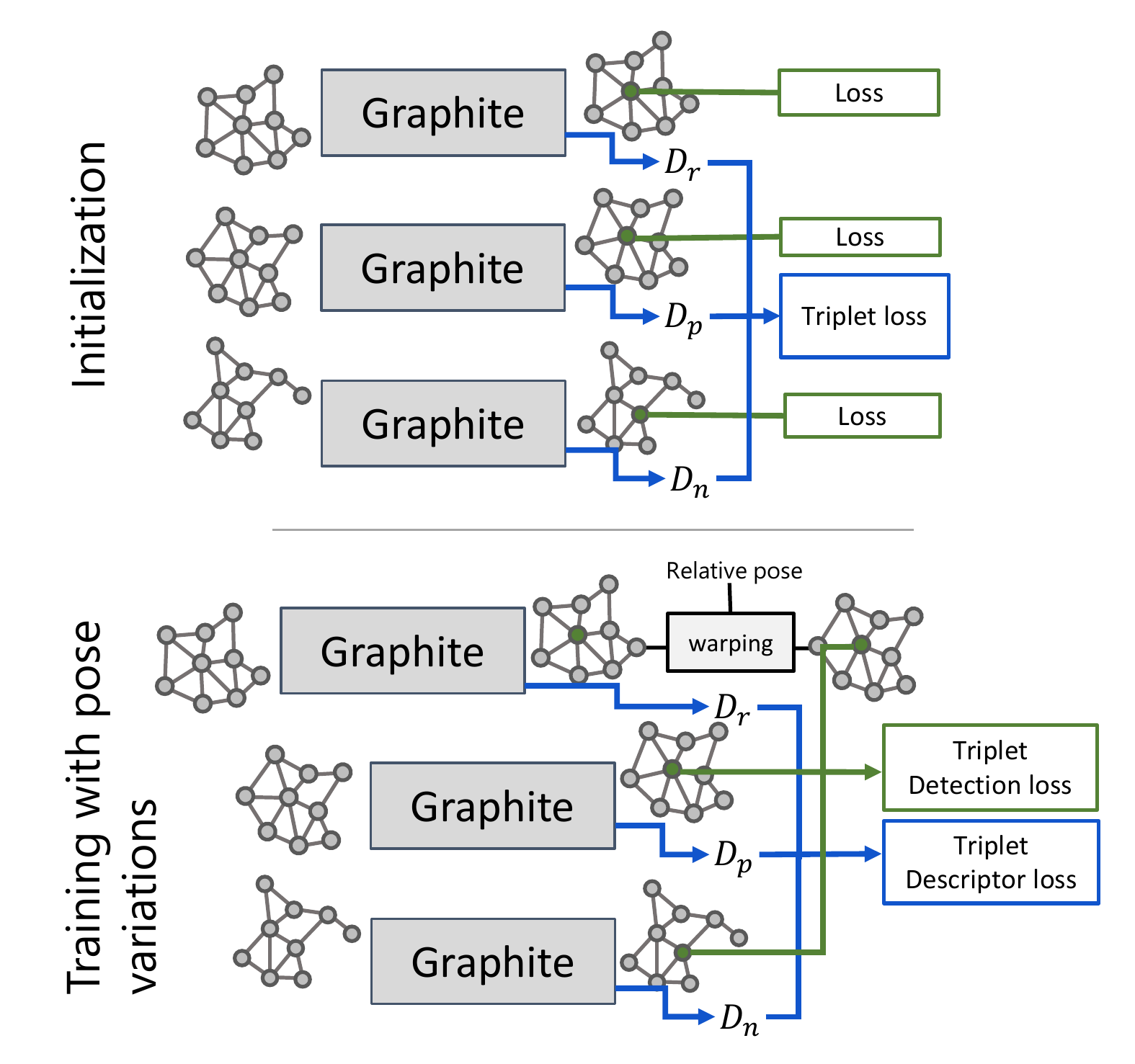}
\caption{Training happens in two stages. In the initialization stage (top), triplet inputs are fed into networks with shared weights. The detectors are learned with supervised loss and the descriptors are learned through triplet loss. (Bottom) we train using 6D pose with minimum supervision. Both detector and descriptor are trained with triplets. In order to measure similarities on point cloud values we warp one point cloud to another using known relative pose.}
\label{fig:stages}
\end{figure}

\subsection{Training with pose variations}
\label{posetraining}

After initializing our model on homemade synthetic primitives dataset, we subsequently fine-tune our feature detector with the support of pose annotations on a more variety of geometries and samplings. Descriptors should be rotational and sampling invariant. If we transform and resample a point cloud with a 6D pose, it should not produce a different descriptor or find a different keypoint. Moreover the salient regions should be consistent under different patch samplings to improve matching. This is particularly significant in real applications e.g. when capturing range data, sampling a surface would produce different point clouds with different resolution and noise. To build a robust descriptor and detector, we use the Graphite model to describe the same surface with varied samplings and poses. Similarly, we remove the direct supervision of value estimation to let the network detect optimal and persistent keypoints from different views using metric learning and triplet loss similar to Eq. \ref{eq:triplet_loss}.

\subsection{Point cloud warping}
Given two point clouds from two different viewpoints with known relative pose, we first tessellate each cloud into small sets of points (patches). In presence of variable sampling rate or noise perturbations transformed points would not lay exactly on corresponding points but rather somewhere on the surface. To facilitate metric learning across different poses we need to warp each patch into one another. With warping, each embedding from the original point set finds its equivalent embedding in the other set to compare.

Lets assume we have a patch $C_i$ from cloud $P$ and we know pose $T_{pq}$ to transform point cloud $P$ to $Q$. Applying known transformation $T$ to $P$ will move us to $P'$, $T_{pq}P=P'$. We find the corresponding patch of $C_i$, $C_j$ in $Q$ using nearest neighbors of $c'$ in $P'$ with
\begin{equation}
   C_j= \{c \in Q :  \exists c' \in P' , c \in knn(Q,c') \}.
\end{equation}
Where $knn(Q,c')$ find k nearest neibors of $c'$ in $Q$. We finally utilize k-NN again to find the weight combinations for corresponding value assignments. The weights are defined as inverse 3d distance in the k nearest neighbors for value warping. 

%% file: sections/Experiments.tex
\section{Experiments and Evaluations}
\subsection{Implementation details}
Our implementation is done with PyTorch Geometric \cite{Fey/Lenssen/2019} and PyTorch \cite{NEURIPS2019_9015} for graph definition and graph convolutions. To process, sample and operate with point clouds, we leverage the Open3D library \cite{Zhou2018}. The evaluation follows the example implementation provided by \cite{GojcicTheDensities} and \cite{wang2019deep}.
We train and evaluate all pipelines on an Intel Core i7-8700K CPU 3.70GHz $\times$ 12 and an Nvidia GeForce RTX 2080 Ti GPU. On this hardware average Graphite calculation for 100 patches takes 0.018 seconds on GPU. 

\subsection{Synthetic Primitive Corners}
\label{syn_dataset}
As introduced in Section \ref{init} and inspired by the 2D processing of MagicPoint \cite{detone2017toward}, we start with synthetic data training with the aim to guide the network in order to learn different shapes and primitive differential geometric concepts to locate corners as keypoints. We create a rendering pipeline with a pair of cameras at random poses pointing towards a shape corner from the object. The corner of focus can associate 3-10 faces with varying heights and curvatures. Every instance is then rendered from the camera pairs to simulate different point sampling as it would occur in a natural scene. The depth maps are then back-projected to a point cloud given the known camera intrinsics. In total, we produce 20k random patch pairs. Each pair is grouped with a random instance which together form a triplet. A random patch of size $n$ is sampled in the vicinity of the corner. Every patch is then converted as explained in Section \ref{graphdef} with a fixed radius $r$, and the nodes are annotated with a value $Y$ inverse to the length of their path to a target keypoint node. The target node is valued with 1. In the supplementary material you can find sample triplets from this dataset .

\begin{table}[t]
\begin{center}
\footnotesize
\begin{tabular}{l |p{0.15cm}r r r}
\hline
Method & & MSE & RMSE & MAE \\
\hline\hline
\multirow{2}{*}{ICP} & R & 892.60 & 29.88 & 23.63\\
& t &  8.60 & 2.93 & 2.52 \\ 
\hline
\multirow{2}{*}{Go-ICP} & R & 192.26 & 13.87 & 2.91\\
& t & 0.05 & 0.22 & 0.06 \\
\hline
\multirow{2}{*}{FGR} & R &  97.00 & 9.85 & 1.45 \\
& t &  0.02 & 0.14 & 0.02 \\ 
\hline
\multirow{2}{*}{PointNetLK} & R & 306.32 & 17.50 & 5.28\\
& t & 0.08 & 0.28 & 0.07 \\ 
\hline
\multirow{2}{*}{DCP-v1} & R &  19.20 & 4.38 & 2.68\\
& t &  \textbf{$<$0.01} & 0.05 & 0.04 \\ 
\hline
\multirow{2}{*}{DCP-v2} & R & 9.92 & 3.15 & 2.01\\
& t &   \textbf{$<$0.01} & \textbf{0.05} & \textbf{0.03} \\ 
\hline
\multirow{2}{*}{\textbf{Ours}} & R & 7.44 & 2.73 & 1.49\\
& t &  0.31 & 0.56 & 0.38 \\ 
\hline
\multirow{2}{*}{\textbf{Ours + ICP}} & R & \textbf{0.75} & \textbf{0.86} & \textbf{0.11}\\
& t &  0.09 & 0.30 & 0.07 \\ 
\hline
\end{tabular}
\end{center}
\caption{Point cloud registration comparison on ModelNet objects from unseen categories.\label{table:modelnet}}
\end{table}

\subsection{ModelNet Object Registration}
Model40 \cite{Wu3DShapes} is a dataset consisting of 3D meshes in 40 different categories. For each category it contains synthetic CAD models. We sample random point clouds uniformly from the CAD model. For a fair comparison we follow the repository of \cite{wang2019deep} to sample 1024 points on the surface. For registration based on ModelNet objects, we randomly generate a rotation and translation (pose) and apply it to the source point cloud as in \cite{wang2019deep}. A random permutation is applied to the resulting list of target points. We then generate uniformly distributed patches with random seeds on three different scales of the point cloud to learn robust descriptors across different scales. It is worth mentioning that patches may not have a (fully) corresponding patch on the other point cloud. Each patch is then converted to a graph and sets of graphs are stored for each pose. We normalize the cloud into unit cube (1m) and apply perturbation augmentation by applying Gaussian noise ($\sigma=0.1cm$) on point coordinates and normals. For both training and testing, we generate 70 patches per object to describe.

\begin{table}[t]
\begin{center}
\footnotesize
\begin{tabular}{l| p{0.15cm}r r r}
\hline
Method & & MSE & RMSE & MAE \\
\hline\hline
\multirow{2}{*}{ICP} & R & 882.56 & 29.71 & 23.56\\
& t & 8.45 & 2.91 & 2.49 \\ 
\hline 
\multirow{2}{*}{Go-ICP} & R & 131.18 & 11.45 & 2.53\\
& t & 0.05 & 0.23 & 0.042 \\
\hline

\multirow{2}{*}{FGR} & R & 607.69 & 24.65 & 10.06\\
& t &  1.19 & 1.09 & 0.27 \\ 
\hline
\multirow{2}{*}{PointNetLK} & R &  256.16 & 16.00 & 4.60\\
& t & 0.047 & 0.216 & 0.057 \\ 
\hline
\multirow{2}{*}{DCP-v1} & R & 6.93 & 2.63 & 1.52\\
& t &   \textbf{$<$0.01} & \textbf{0.02} & 0.02 \\ 
\hline
\multirow{2}{*}{DCP-v2} & R & \textbf{1.17} & \textbf{1.08} & 0.74\\
& t &  \textbf{$<$0.01} & \textbf{0.02} & \textbf{0.01} \\ 
\hline
\multirow{2}{*}{\textbf{Ours}} & R & 17.76 & 4.21 & 2.35\\
& t &  0.39 & 0.63 & 0.44 \\ 
\hline  
\multirow{2}{*}{\textbf{Ours + ICP}} & R & 4.03 & 2.01 & \textbf{0.31}\\
& t &  0.10 & 0.31 & 0.08 \\ 
\hline
\end{tabular}
\end{center}
\caption{Effect of Gaussian noise on point cloud registration on ModelNet40 dataset\label{table:guassian}
}
\end{table}

\begin{figure}
\centering
\includegraphics[width=\linewidth]{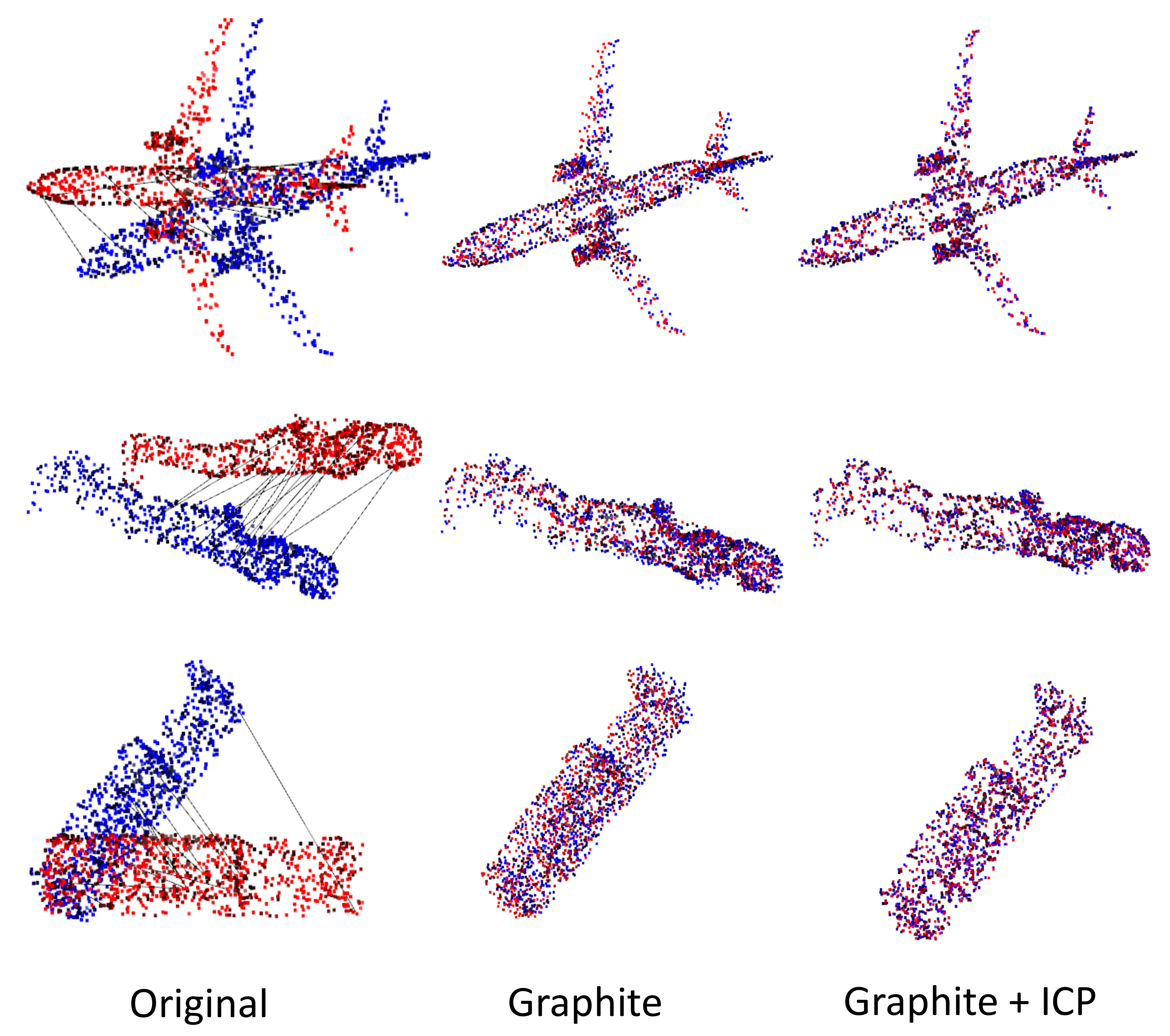}
\caption{Qualitative evaluation on ModelNet40  \cite{Wu3DShapes} dataset registration in presence of noise. Graphite matching followed by RANSAC-based pose estimation provide an almost perfect initial pose for follow-up ICP refinement.}
\label{fig:res_modelnet}
\end{figure}

\begin{figure}
\centering
\includegraphics[width=\linewidth]{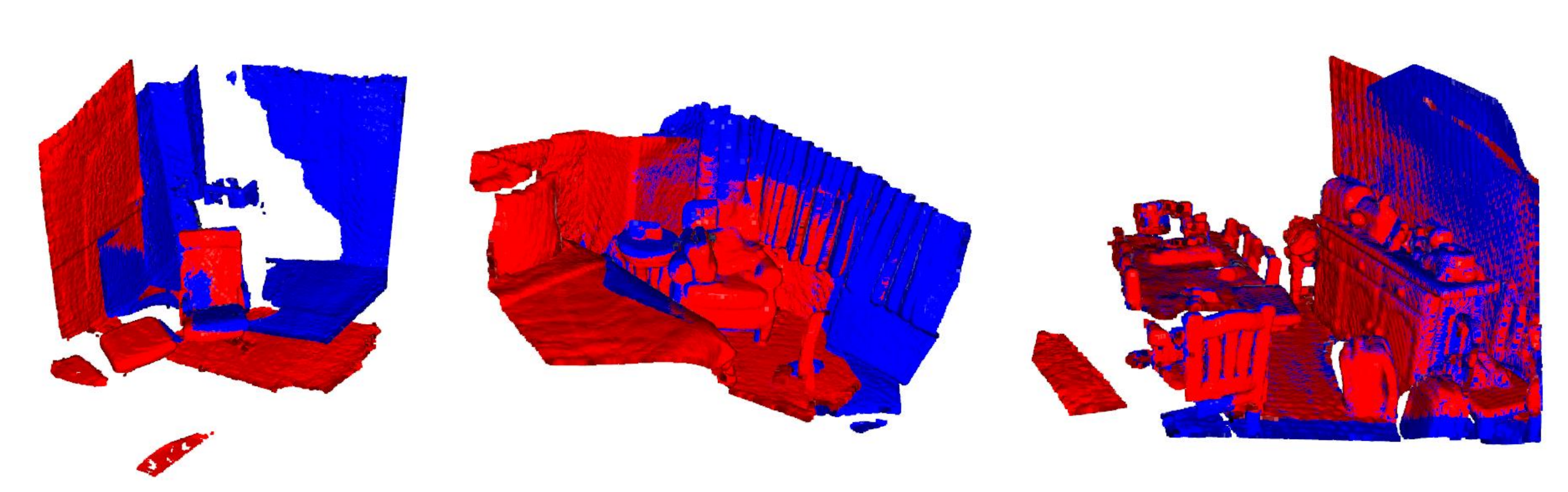}
\caption{Qualitative evaluation on 3DMatch \cite{Zeng3DMatch:Reconstructions} dataset registration. Random registration examples from three scenes MIT, Home 1 and Hotel 1.}
\label{fig:res_3dmatch}
\end{figure}

\begin{table*}[t]
\begin{center}
\footnotesize
\begin{tabular}{l|c c|c c c c c |c c}
\hline
&\multicolumn{2}{c|}{Handcrafted} & \multicolumn{5}{c|}{Trained on 3DMatch} & \\
Method & FPFH & SHOT & 3DMatch &  PPFNet & PPF-FoldNet & Perfect-Match & Perfect-Match & CGF & Graphite \\
& (33 dim) & (352 dim) & (512 dim)  & (64 dim) & (512 dim) & (16 dim) & (32 dim) & (32 dim) & (32 dim)\\
\hline\hline
Kitchen & 43.1 & 74.3 & 58.3 &  89.7 & 78.7 & 93.1 & 97.0 & 60.3 & 64.82\\
Home 1 & 66.7 & 80.1 & 72.4 & 55.8 & 76.3 & 93.6 & 95.5 &  71.1 & 83.97\\
Home 2 & 56.3 & 70.7 & 61.5 & 59.1 & 61.5 & 86.5 & 89.4 &  56.7 & 68.26\\
Hotel 1 & 60.6 & 77.4 & 54.9 &  58.0 & 68.1 & 95.6 & 96.5 & 57.1 & 80.77\\
Hotel 2 & 56.7 & 72.1 & 48.1 & 57.7 & 71.2 & 90.4 & 93.3 & 53.8 & 80.53\\
Hotel 3 & 70.4 & 85.2 & 61.1 & 61.1 & 94.4 & 98.2 & 98.2 & 83.3 & 96.29\\
Study & 39.4 & 64.0 & 51.7 & 53.4 & 62.0 & 92.8 & 94.5 & 37.7 & 82.53\\
MIT Lab & 41.6 & 62.3 & 50.7 & 63.6 & 62.3 & 92.2 & 93.5 &  45.5 & 76.62\\
\hline
Average & 54.3 & 73.3 & 57.3 & 62.3 & 71.8 & 92.8 & 94.7 & 58.2 & 79.22\\
\hline
STD & 11.8 & 7.7 & 7.8 & 11.5 & 9.9 & 3.4 & 2.7 & 14.2 &  9.10 \\
\hline
\end{tabular}
\end{center}
\caption{Results on 3DMatch Geometric Registration Benchmark.\label{table:3dmatch}}
\end{table*}

We compare our registration performance on the ModelNet40 \cite{Wu3DShapes} dataset based on the evaluation criteria provided in \cite{wang2019deep}. We train our model with the first 20 categories and evaluate with the 20 unseen categories. We first calculate Graphite features and keypoints per patch in each frame and then find correspondences based on Euclidean feature distance. Pairs of matched keypoints from our detected pool are then used to calculate a pose with an SVD-based pose estimation. We then calculate the Mean Average Error (MAE), Mean Squared Error (MSE) and Root Mean Squared Error (RMSE) on each rotation (in degrees) and translation (in cm) component. Table \ref{table:modelnet} compares our registration errors with state of the art registration methods on this dataset.\\
Methods such as ICP, Go-ICP and DCP are iterative optimization methods. Their convergence is sensitive to the initialization. Therefore, the translation components is minimized comparably well based on the centroid of the full cloud, while the rotation estimate is not very robust in cases of minor overlap or for partial scans. We also add ICP as a consecutive refinement stage after calculating our pose. We reach state of the art rotation error with a significant margin on all estimated metrics.

\subsection{Registration under Noise }
We also study the robustness of our pipeline in presence of noise. We add Gaussian noise to the target point cloud coordinates with a standard deviation of 1~cm. Similar to \cite{wang2019deep}, we test our approach with unseen test instances from all trained categories. We detect local keypoints per patch and measure descriptors to match them. We use an SVD based solver to predict the pose. Table \ref{table:guassian} shows error results in comparison to the state of the art method on the ModelNet40 dataset. While the squared errors reflect some minor outliers, we keep being the method with best performance on MAE error with 0.31 degrees. Figure \ref{fig:res_modelnet} shows sample registration results on ModelNet40\cite{Wu3DShapes} dataset. For extensive evaluations you can refer to supplementary material.

\subsection{3DMatch Descriptor Matching}
The 3DMatch benchmark \cite{Zeng3DMatch:Reconstructions} is a 3D descriptor and geometric registration benchmark. It consists of 7 indoor scenes with multiple point cloud frames each. The point cloud instances are partial views of a fixed scene captured with an RGB-D sensors. We evaluate on this benchmark to assess the real-world applicability of Graphite. For evaluation, we follow the official repositories of \cite{GojcicTheDensities,Zeng3DMatch:Reconstructions} where 5k keypoint coordinates are provided per frame. As the scans include a huge set of points, we first sub-sample points with voxel based sub-sampling and then form patches around the list of given seed coordinates. We form patches with $n=225$ points to describe the same vicinity used to describe other features.\\

 Contrary to the state of the art methods such as \cite{Zeng3DMatch:Reconstructions,Deng2018PPF-FoldNet:Descriptors,Deng2018PPFNet:Matching,GojcicTheDensities} which have trained their models on 3dMatch data or other realistic scans, we have trained our model on synthetic point clouds only. This test demonstrates the transfer and generalization capabilities of our method applied on real data registration task.\\
Given the pool of stored locations from 3dMatch\cite{Zeng3DMatch:Reconstructions}, we describe each local patch and perform matching with our descriptors, we then match them based on their Euclidean distance and perform RANSAC based registration. We take the same RANSAC iteration and settings as used in 3DSmoothNet \cite{GojcicTheDensities}. We calculate recall values instructed in  \cite{Deng2018PPF-FoldNet:Descriptors,GojcicTheDensities} with $\tau_1=0.1m$ and $\tau_2=0.05$.\\
In Table \ref{table:3dmatch} we present our results for the 3DMatch benchmark. We achieve a high recall rate in most of the scenes while having a dense representation (with 32 dimension) and a super lightweight model. In Figure \ref{fig:res_3dmatch} some example registrations drawn from the benchmark fragments are shown. Graphite demonstrate satisfying results even in cases with very small overlap. Moreover, in contrast to Perfect Match \cite{GojcicTheDensities}, we do not use a memory-hungry  voxelization representation, but rely on computationally more efficient graph operations on point clouds through simple matrix multiplications presented in Eq. \ref{eq:tag}. For extensive qualitative evaluation of 3DMatch we refer the interested reader to our supplementary material.

%% file: sections/Conclusion.tex
\section{Conclusion}
We propose A lightweight patch descriptor which can represent point clouds ideal for expensive problems. Our graph-based model efficiently learns shape features and can detect salient keypoints given synthetic prior training followed by self-supervised metric learning. The extracted keypoints alongside the condensed descriptor can be used in the task of point cloud registration. We improve the state of art on object point cloud registration and prove solid performance and generalization on real indoor scans. Graphite can enable fast computation and dense representation of point clouds for modern 3D vision problems, replacing classical descriptors and sampling techniques.

%% file: sections/supp/supp_1.tex
\section{Synthetic Primitive Corners Dataset}
\begin{figure*}
\centering
\includegraphics[clip, trim=0.5cm 3cm 0.5cm 3cm, width=\linewidth]{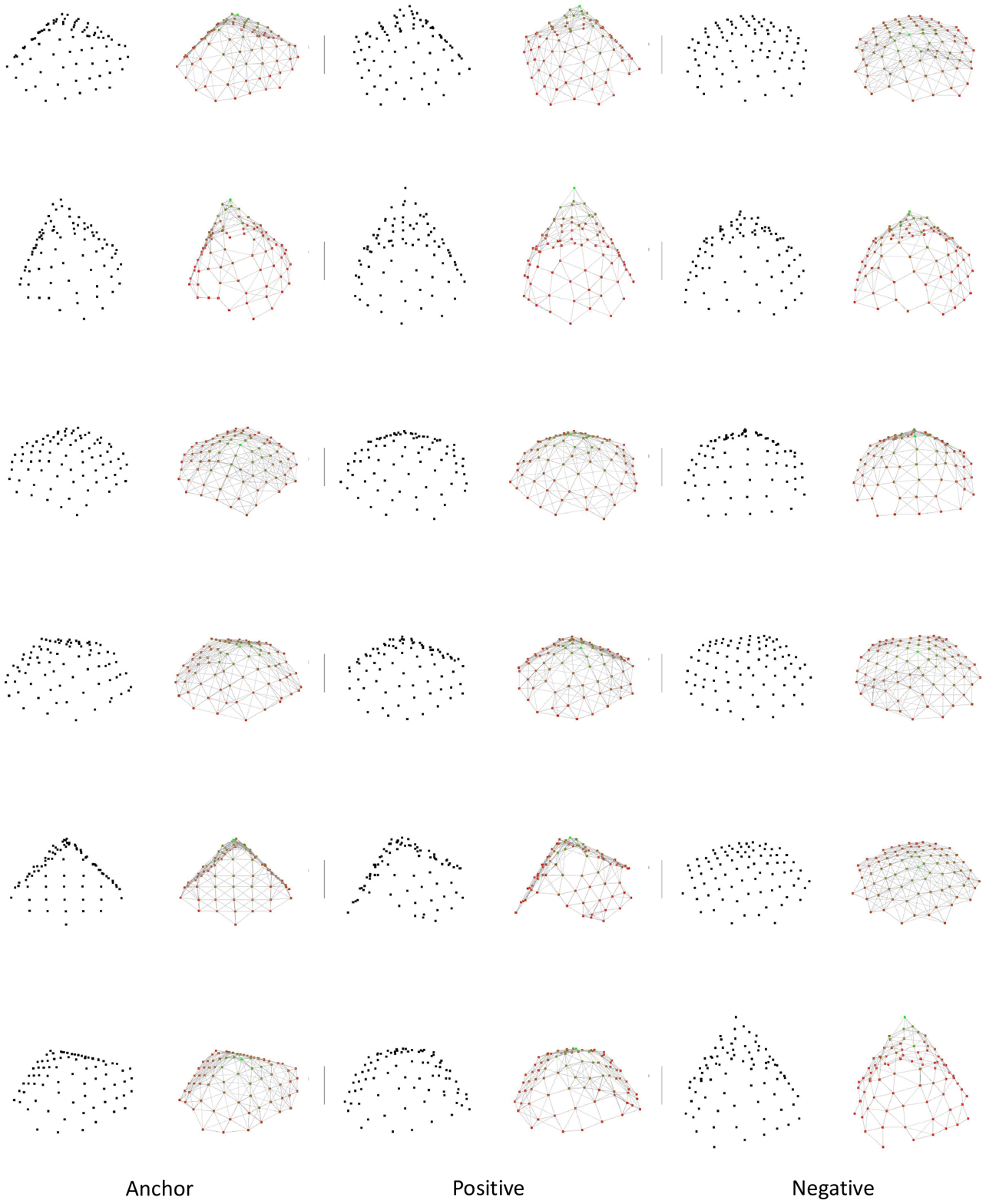}
\caption{Sample triplets drawn from our synthetic primitive corners dataset and their corresponding graphs. The edge weights are visualized using color intensities of the lines. The further the nodes are located the lower the edge between them is weighted. Moreover, the node values $V_i$ for supervision are depicted using a color map (green: high value, red: low value) }
\label{fig:corner_graphs}
\end{figure*}

In this section, we demonstrate our synthetic dataset introduced in section 4.2 of the paper which is used to pretrain our network with simple shape corners. Random shape corners are rendered into depth frames and back-projected to form point cloud patches. The patches are sub-sampled using voxel-based sub-sampling to capture sufficient area of the shape for description. The points are then normalized to unit cube, transformed into the local reference frame, and converted to graphs by finding connecting edges between neighboring nodes as explained in the paper. Figure \ref{fig:corner_graphs} shows triplet samples from our dataset and their representative graphs for $n=81$. Green areas show high-value regions near corners which potentially possess a keypoint.

%% file: sections/supp/supp_2.tex
\section{ModelNet40 Registeration}
\begin{figure*}
\centering
\includegraphics[clip, trim=0.5cm 3cm 0.5cm 3cm, width=\linewidth]{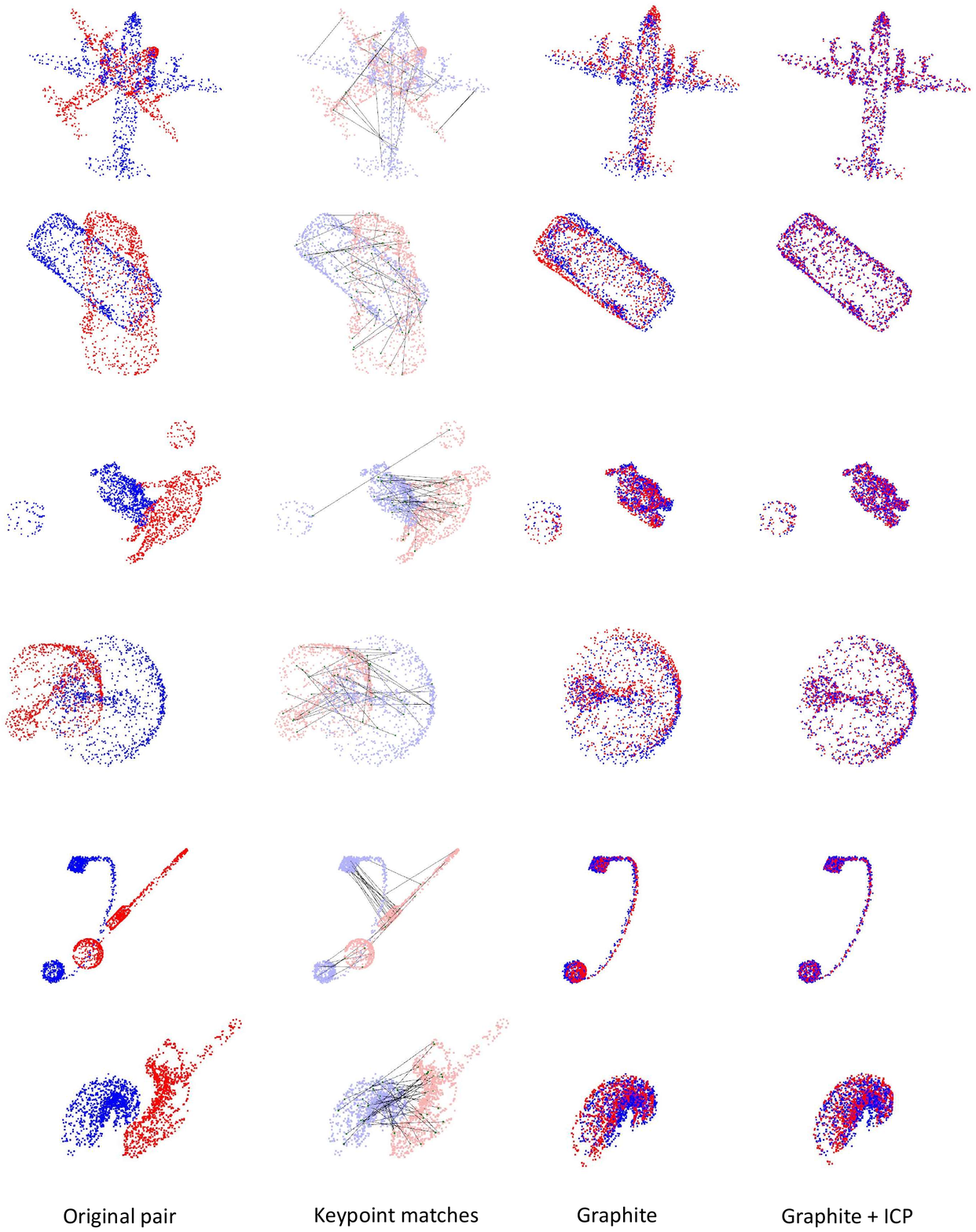}
\caption{Sample registration results on ModelNet40 \cite{Wu3DShapes} with unseen instances and added noise perturbations. We demonstrate that our keypoints extracted from randomly sampled patches can be matched properly using the proposed descriptor to find a good registration. Following Graphite-based registration, an ICP refinement stage can fine-tune the estimated 6D pose as seen on the right.  }
\label{fig:res_modelnet_ext}
\end{figure*}

As explained in the paper, we train and test our model on ModelNet40 \cite{Wu3DShapes} patches. Similar to \cite{wang2019deep} we create a randomly sampled point cloud of size 1024 from each object instance and transform a copy of the instance to a random pose to create a registration pair. The pairs are then converted to patches and each patch is converted to a graph and fed to our model to predict descriptors and detect keypoints. We use the descriptor to search for corresponding keypoints and perform an SVD based pose estimation combined with RANSAC to predict the pose. We further refine our pose with ICP registration. Figure \ref{fig:res_modelnet_ext} shows several registration results of instances drawn from the test data with added noise.

%% file: sections/supp/supp_3.tex
\section{3DMatch Descriptor and Detector}
\begin{figure*}
\centering
\includegraphics[clip, trim=0.5cm 6cm 0.5cm 4cm, width=\linewidth]{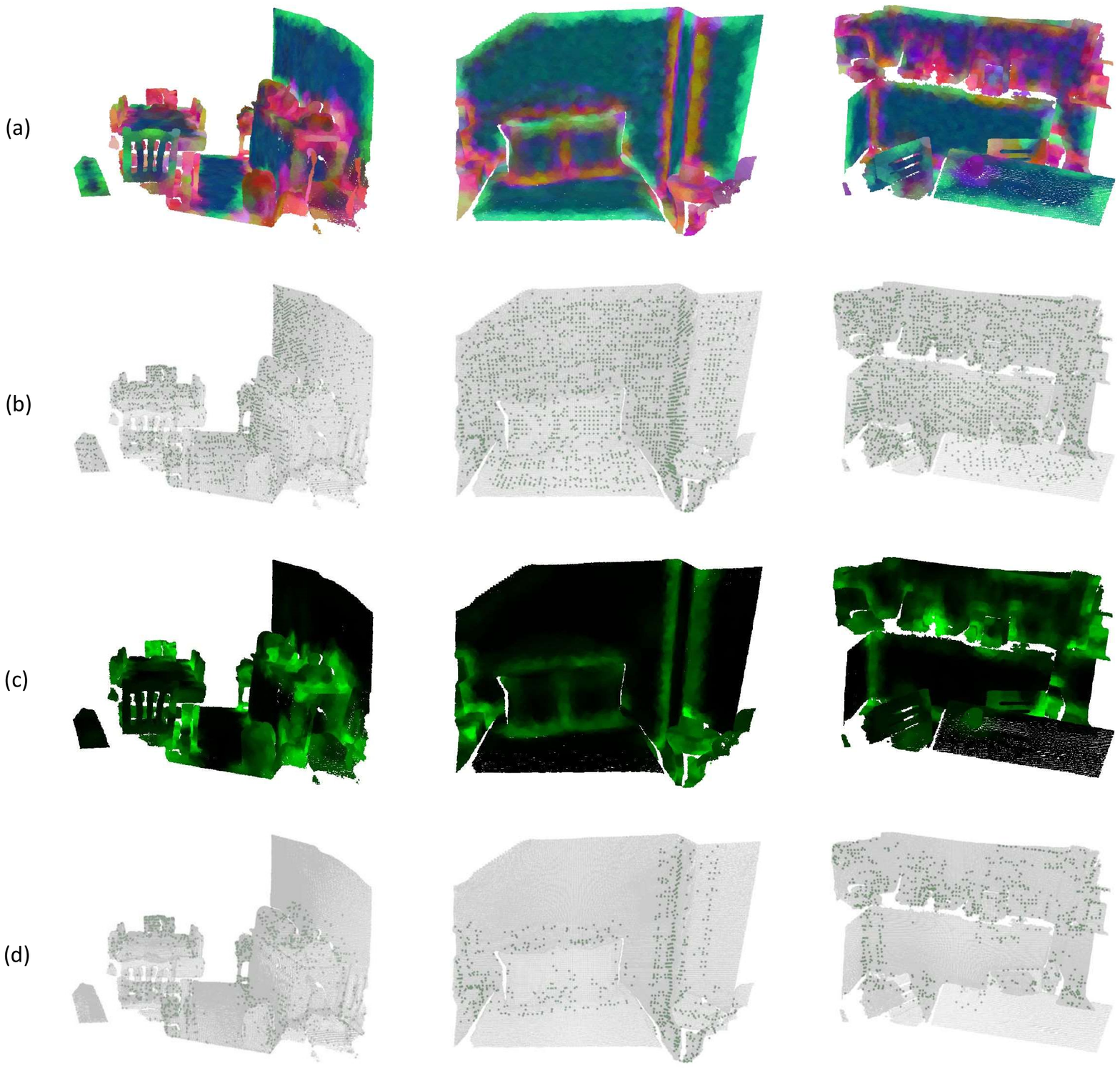}
\caption{ The visualization of descriptor and keypoint detection on unseen 3DMatch dataset \cite{Zeng3DMatch:Reconstructions}. (a) shows descriptor mapped to color space using PCA. (b) shows the fixed seed points suggested by 3DMatch and used for benchmark evaluation. (c) shows 3D keypoint score map $S$ while (d) shows remaining keypoints after validation using the scores.}
\label{fig:3dmatch_vis}
\end{figure*}
In this section we show the performance and generalization of Graphite through qualitative evaluation on an unseen dataset. 3DMatch \cite{Zeng3DMatch:Reconstructions} consists of point cloud fragments of indoor environment scans. We first show how our descriptor sees the patches in the clouds. In order to do that, we reduce our predicted descriptor from ($l=32$) to a lower dimension ($l_{red}=3$) using Principal Component Analysis (PCA). We then Normalize and map the features to RGB color space. The visualization on figure \ref{fig:3dmatch_vis} (a) shows response of our descriptor on different surface structures.\\
Furthermore, we showcase our keypoint detector and scores visually. Figure \ref{fig:3dmatch_vis} (b) shows seed interest points proposed by 3DMatch which are distributed uniformly on the point cloud. Although Graphite is capable of detecting its own keypoints, for a fair comparison with the state of the art, we have utilised the same seed points and the same radius (0.15m) used in \cite{GojcicTheDensities}. As previous works suggest dense uniform seed point generation with 5k or 2k seed points for description, they manage to perform well for registration with the combination of RANSAC. Figure  \ref{fig:3dmatch_vis} (c) shows our keypoint scoring $S_i$ visualisation, where greener areas show more salient keypoint regions, and darker areas show lower scores. This map can be simply used for keypoint validation. Figure \ref{fig:3dmatch_vis} (d) shows our keypoints that are validated using a threshold ($S>0.2$). As observable in the figure, most of the keypoints on the flat regions are invalidated due to a low shape complexity. By means of this filtering, a much relaxed and therefore faster RANSAC can be used for registration. 

%% file: sections/supp/supp_4.tex
\section{3DMatch Registeration}
\begin{figure*}
\centering
\includegraphics[clip, trim=0.5cm 1.5cm 0.5cm 1cm, height=22.5cm]{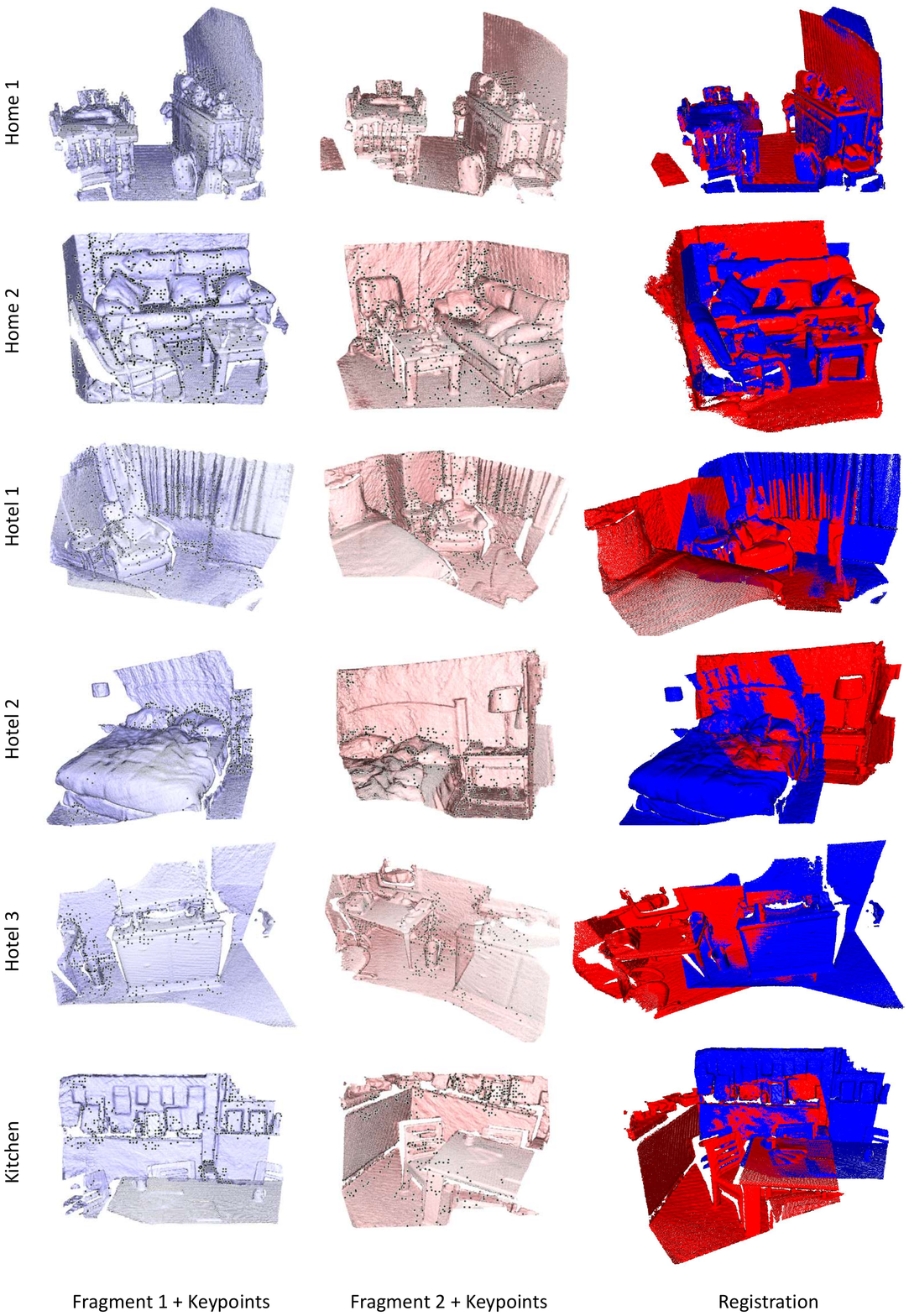}
\caption{ Keypoints and Registration results on 3DMatch \cite{Zeng3DMatch:Reconstructions} benchmark.}
\label{fig:3dmatch_reg_ext}
\end{figure*}
Finally as described in section 4.5 of the paper, we benchmark on 3DMatch Geometric Registration Benchmark \cite{Zeng3DMatch:Reconstructions}. We use the keypoints and descriptors to register two frames of a scene. On average we have ~536 points after descriptor matching for pose estimation. Sample results can be seen in Figure \ref{fig:3dmatch_reg_ext} with detected keypoints, and registered patches. The results illustrates our robust feature-based registration works well under partial views and limited point cloud overlaps. We demonstrate sample results from 6 different scenes to further prove generalization under different settings.